\documentclass[conference]{IEEEtran}
\IEEEoverridecommandlockouts
\usepackage{cite}
\usepackage{amsmath,amssymb,amsfonts}
\usepackage{algorithmic}
\usepackage{graphicx}
\usepackage{textcomp}
\usepackage{xcolor}

\def\BibTeX{{\rm B\kern-.05em{\sc i\kern-.025em b}\kern-.08em
    T\kern-.1667em\lower.7ex\hbox{E}\kern-.125emX}}
\begin{document}

\title{
A Study on the Vulnerability of Test Questions against ChatGPT-based Cheating
}

\author{\IEEEauthorblockN{ Shanker Ram}
\IEEEauthorblockA{\textit{Lynbrook High School} \\
San Jose, USA \\
shankernram@gmail.com}
\and
\IEEEauthorblockN{ Chen Qian}
\IEEEauthorblockA{\textit{Department of Computer Science} \\
\textit{University of California at Santa Cruz}\\
Santa Cruz, California \\
cqian12@ucsc.edu}
}

\maketitle

\begin{abstract}
ChatGPT is a chatbot that can answer text prompts fairly accurately, even performing very well on postgraduate-level questions.
Many educators have found that their take-home or remote tests and exams are vulnerable to ChatGPT-based cheating because students may directly use answers provided by tools like ChatGPT. 
In this paper, we try to provide an answer to an important question: how well ChatGPT can answer test questions and how we can detect whether the questions of a test can be answered correctly by ChatGPT.
We generated ChatGPT’s responses to the MedMCQA dataset, which contains over 10,000 medical school entrance exam questions. We analyzed the responses and uncovered certain types of questions ChatGPT answers more inaccurately than others. In addition, we have created a basic natural language processing model to single out the most vulnerable questions to ChatGPT in a collection of questions or a sample exam. Our tool can be used by test-makers to avoid ChatGPT-vulnerable test questions. 
\end{abstract}


\section{Introduction}
ChatGPT is an advanced chatbot developed by OpenAI, and it is proficient at answering test questions from any educational level. Because of the recent increase in online learning, using ChatGPT to cheat on tests is more prevalent. According to Yahoo Finance \cite{yahoo}, some teachers have caught over 26\%  of students attempting to cheat on tests using ChatGPT. In order to combat AI usage on tests, we conduct this study to find techniques test-makers can use to detect questions that chatbots can answer and create those that are not vulnerable to ChatGPT. 

We first conduct a preliminary study on the accuracy of using ChatGPT to answer existing test questions. 
After initially testing ChatGPT on over 10,000 questions, with the use of the Python ChatGPT API, we found that it answered these medical school entrance exam questions correctly about 60\% of the time. ChatGPT's above average performance on postgraduate-level questions is alarming, given the frequency of cheating using AI on tests in college and high school. \cite{college}. We tried looking at basic features in the data, including question lengths and the number of words per question. However, we soon realized that ChatGPT was advanced enough to answer correctly irrespective of these metrics. Our important discovery is that the true markers of a question’s susceptibility against ChatGPT lies in more obscure features. 

By scouring the data and analyzing the questions ChatGPT got right and wrong, we discovered many helpful similarities in the questions it got right and the ones it got wrong. By applying the techniques we present, test-makers can improve the quality of the questions they create such that ChatGPT cannot answer them accurately. 

We also created an NLP model with a neural network so that given the text of a question, the model would be able to immediately answer whether ChatGPT would answer the question correctly or incorrectly, as well as the likelihood of that outcome happening. We attempted this through the use of Google Colab’s built-in GPU, making use of the PyTorch framework \cite{pytorch} in Python. The finished model correctly answers whether a question is vulnerable to ChatGPT with a 60\% accuracy; in addition, when the model had a confidence score of over 85\% in its answer, which happened about half of the time, its accuracy rose to over 70\%. Overall, this model is a satisfactory tool in at least marginally filtering out GPT-susceptible questions. 

\section{Dataset}

Selecting the correct dataset for this task was important, since we need a dataset with college-level multiple-choice questions, over 10,000 problems, and answers to each of those problems. Additionally, we prefer problems related to life science because processing questions with math notation was tedious and inconsistent in Python. We ended up choosing MedMCQA \cite{MedMCQA}, which has 200,000 multiple-choice medical school entrance exam questions and answers; as an additional bonus, each of the questions has labels with the branch of medicine it tests, providing another feature to later analyze in the results. 

\subsection{Data Collection}

In order to gather the responses from ChatGPT for each of the questions, we used the ChatGPT Python API; we settled on testing about 13000 questions since it took about 2 seconds for the API to process and respond to each prompt. Additionally, it should be noted that ChatGPT is a nondeterministic model \cite{nondeterministic}, meaning the same prompt can be entered multiple times with different responses. This means that not all of the data gathered is necessarily representative of that done by using the ChatGPT model available on the OpenAI website. However, although the Python API often tends to output different responses than the website, the overall accuracies on questions are actually similar for both. To test this, we manually entered a subset of 100 random problems from the MedMCQA dataset into the ChatGPT model on the OpenAI website to test its performance. The same 100 problems were also given to the ChatGPT Python API. The results are shown in Table 1.

\begin{table}[htbp]
\caption{Accuracy for API vs. website}
\begin{center}
\begin{tabular}{ | p{4cm} | p{4cm} |} 
  \hline
  Accuracy on 100 random questions from the dataset (website) & Accuracy on 100 random questions from the dataset (Python API) \\ 
  \hline
  63\% & 63\% \\ 
  \hline
\end{tabular}
\end{center}
\end{table}

Remarkably, both the API and the website got exactly 63 of the 100 questions right, although not the same 63. Regardless, the existing sources of error are extremely small, so the trends present in the data can still be extrapolated and used to our advantage when making test questions.

\section{Apparent Trends in Data}

ChatGPT performs very differently when faced with varying types of questions. In some cases, the differences had no effect, while in others, GPT's accuracy is reduced by over 50\%.

\subsection{Structuring and Complexity of Questions has no Effect on ChatGPT}

When applying basic NLP techniques \cite{basicnlp} toward the questions, including calculating words per question, characters per question, sentences per question, words per sentence, characters per sentence, characters per word, special figures per question, and numerical figures per question, we found that these metrics did not correspond to the problems ChatGPT answered correctly or the problems it answered incorrectly, as shown in Fig. 1. This implies that any basic features will not separate complex problems from simple problems for ChatGPT.

After testing many simple NLP techniques, we moved on to advanced lexical richness metrics. We tried 10 of these metrics \cite{advancednlp}, including the Flesch Reading Ease score, the Flesch Kincaid Grade Level test, the SMOG index readability score, the Coleman-Liau index, the Automated readability index, the Dale-Chall readability score,  Gunning Fog index, and the Gulpease index. Like with the basic NLP techniques, all of the lexical richness metrics were the same independent of whether ChatGPT got the problem right or wrong, as shown in Fig. 2. We also tried comparing the cosine similarity \cite{cosine} between problems ChatGPT got right and wrong, and again, the results were very similar. 

In conclusion, it is not possible to fool ChatGPT by making problems sound more complex or using complicated synonyms of words to try and confuse it. Questions must be changed in different ways to actually have a chance of stopping ChatGPT from correctly answering them.

\begin{figure}
	   \includegraphics[width=0.4\textwidth]{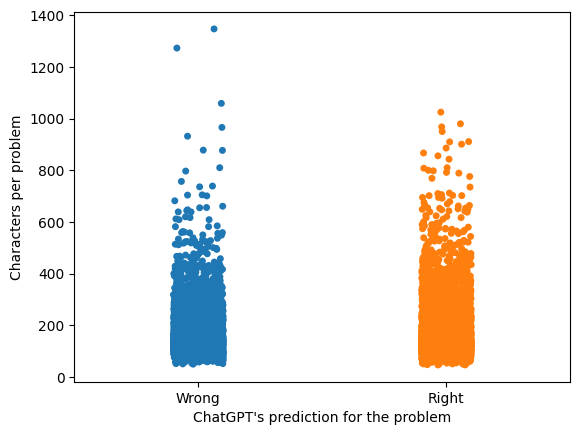}	
	\caption{Categorical scatterplot with slight random jitter to avoid overplotting} 
\end{figure}
\begin{figure}
	   \includegraphics[width=0.4\textwidth]{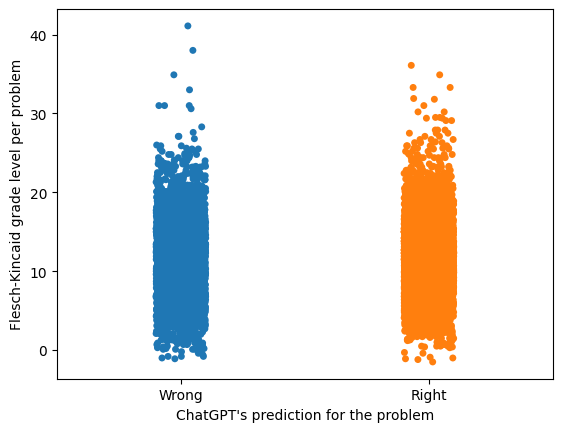}	
	\caption{Categorical scatterplots with slight random jitter to avoid overplotting} 
\end{figure}

\subsection{Using Multi-Select Problems has Little to no Effect on ChatGPT’s Performance}

\begin{figure}
	   \includegraphics[width=0.4\textwidth]{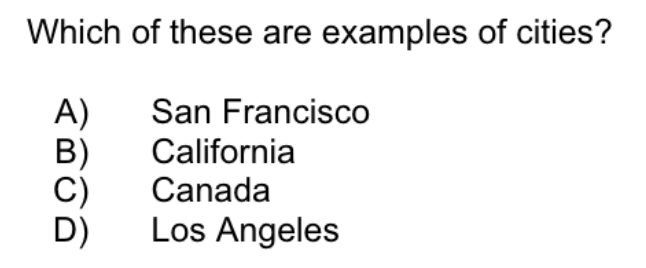}	
	\caption{Example of multi-select problem, the correct answer is A and D} 
\end{figure}

We assumed ChatGPT would perform worse on multi-select problems, an example of which is shown in Fig. 3, and it did perform slightly worse, dropping 4-5\% in accuracy on these types of questions. Still, this does not justify using multi-select problems over other types of problems to combat ChatGPT since

a) While ChatGPT is worse at answering these types of questions, so are test takers.

b) Test-makers would need to spend much more time creating these questions instead of normal MCQ.

c) The relatively small drop is not large enough to completely rely on these types of questions.

\subsection{Adding Extra Option Choices has no Effect on ChatGPT’s Performance}

Adding extra option choices would presumably create a much larger challenge \cite{extraoptions}, but to our surprise, ChatGPT had almost the same accuracy on questions regardless of the number of answer choices. When adding a fifth answer choice to MCQ, ChatGPT’s accuracy was just slightly lower, decreasing by about 2\%.

When analyzing the responses provided to these questions, it is easy to see that the reason for this phenomenon is that ChatGPT treats all questions like free-response questions, where it does not even consider the answer choices until selecting its final answer. In fact, the only case where the number of answer choices affects ChatGPT is when it has to guess, where the probability of a guess being correct is higher with only four answer choices. This explains the 2\% drop seen earlier. 

The good news for question-makers is that they do not have to worry about trying to make free-response questions \cite{frqhard} to combat ChatGPT, since ChatGPT answers both MCQ and FRQ with relatively equal accuracy.

\subsection{ChatGPT Struggles More with Questions with the Word ``except'' in Them}

To be more specific, questions in the dataset with the word ``except'' in them typically had the phrase “All of the following are true except”. An example of what such a problem could look like is shown in Fig. 4.

Interestingly, ChatGPT performed about 6\% worse on questions worded similarly to the example above. There are many possibilities for why this could be the case but after analyzing ChatGPT’s responses to questions like this, one theory is that for normal questions, ChatGPT has to verify that one statement is true; in “except” questions, it has to verify that three statements are true. This is a potential reason for ChatGPT’s worse performance.

However, it must be noted that 6\% is not that large of a gap. On a 50-question test, applying this technique could save about 3 questions from being answered accurately by ChatGPT, but this is still not a huge impact. 

\begin{figure}
	   \includegraphics[width=0.4\textwidth]{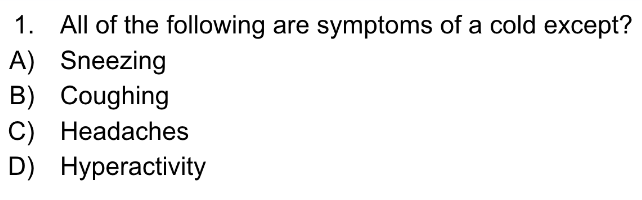}	
	\caption{Example of problem with "except" in it} 
\end{figure}

\subsection{A Major Indicator of ChatGPT’s Success at Answering a Question Correctly is the Topic the Question is Based On}

As expected, the topic of the question ChatGPT responds to is a huge factor in whether it gets the question right or wrong. As it turns out, ChatGPT is very good at answering questions with topics like biochemistry or psychiatry, with accuracy rates over 70\%, and it is much worse at answering questions with topics like dentistry, with accuracy rates under 50\% for these types of questions.

A simple theory as to why ChatGPT is worse at answering certain questions comes from the topics themselves. Since ChatGPT was trained off of data from the Internet, it can be assumed it would be more proficient at answering questions related to more common topics. As information related to biochemistry and psychiatry is more readily available \cite{commonmajors} on the Internet in comparison to dentistry, there could have been less training data for ChatGPT to train on dentistry. A full list of accuracies for each topic is shown in Table 2. 

\begin{table}[htbp]
\caption{Topic of Question vs. Accuracy}
\begin{center}
\begin{tabular}{ | p{4cm} | p{4cm} | } 

  \hline
  \textbf{Topic} & \textbf{ChatGPT’s accuracy in answering questions with this topic} (\%) \\ 
  \hline
  Anatomy & 54  \\ 
  \hline
  Biochemistry & 71  \\ 
  \hline
  Surgery & 59  \\ 
  \hline
  Ophthalmology & 55  \\ 
  \hline
  Physiology & 63  \\ 
  \hline
  Social and Preventive Medicine  & 58  \\ 
  \hline
  Gynecology and Obstetrics & 56  \\ 
  \hline
  Anesthesia & 51  \\ 
  \hline
  Psychiatry & 72  \\ 
  \hline
  Microbiology & 64  \\ 
  \hline
  Medicine & 60  \\ 
  \hline
  Pharmacology & 70  \\ 
  \hline
  Dental & 48  \\ 
  \hline
  EMT & 55  \\ 
  \hline
  Forensic Medicine & 50  \\ 
  \hline
  Pediatrics & 58  \\ 
  \hline
  Orthopedics & 53  \\ 
  \hline
  Radiology & 52  \\ 
  \hline
  Pathology & 64  \\ 
  \hline
  Skin & 67  \\ 
  \hline
\end{tabular}
\end{center}
\end{table}

\subsection{ChatGPT Drastically Overpredicts the Options “All of the above” and “None of the above”}

When first running the code on questions with “None of the above” and “All of the above” as possible answer choices, we found that ChatGPT is very good at answering questions with “All of the above” as an answer choice and it is very poor at answering “None of the above” as an answer choice. However, this data did not tell the full story. 

When looking at the data for questions with the answer choice “None of the above” again, we split these questions up into two categories—questions where the correct answer was “None of the above”, and questions where the correct answer was not “None of the above”. For the questions where the correct answer \textit{was} “None of the above”, ChatGPT’s accuracy was a high 77\%, and for the questions where “None of the above” was incorrect, ChatGPT’s accuracy was a shockingly low 48\%. This discrepancy is shown in Table 3.

\begin{table}[htbp]
\caption{Accuracy vs. presence of "None of the above" option}
\begin{center}
\begin{tabular}{ | p{2.5cm} | p{2.5cm} | p{2.5cm} |} 
  \hline
  ChatGPT’s accuracy on questions where “None of the above” is the right answer & ChatGPT’s accuracy on questions where “None of the above” is not the right answer & ChatGPT’s accuracy on questions without “None of the above” as an answer choice \\ 
  \hline
  78\% & 47\% & 59.5\%\\ 
  \hline
\end{tabular}
\end{center}
\end{table}

When repeating the same process for questions with the answer choice “All of the above”, we found similar results, but exaggerated even further. For the questions where the correct answer \textit{was} “All of the above”, ChatGPT’s accuracy was an incredibly high 98\%, and for the questions where “All of the above” was incorrect, ChatGPT’s accuracy was much lower at 16\%. This discrepancy is shown in Table 4.

\begin{table}[htbp]
\caption{Accuracy vs. presence of "All of the above" option}
\begin{center}
\begin{tabular}{ | p{2.5cm} | p{2.5cm} | p{2.5cm} |} 
  \hline
  ChatGPT’s accuracy on questions where “All of the above” is the right answer & ChatGPT’s accuracy on questions where “All of the above” is not the right answer & ChatGPT’s accuracy on questions without “All of the above” as an answer choice \\ 
  \hline
  98\% & 16\% & 59.5\%\\ 
  \hline
\end{tabular}
\end{center}
\end{table}

This huge discrepancy in accuracy can be explained by ChatGPT’s tendency to select “None of the above” or "All of the above" even when it is not the right answer. It seems to overpredict these options as the answer in these cases. Table 5 illustrates this tendency. For example, when option choice "D" is "None/All of the above" and it is wrong, ChatGPT still selects "D" 81\% of the time. 

\begin{table}[htbp]
\caption{Accuracy based on whether "(D) None/All of the Above" was correct}
\begin{center}
\begin{tabular}{ | p{1.5cm} | p{3cm} | p{3cm} |} 
  \hline
  Phrase & Percentage of ChatGPT's responses which are “(D) None/All of the Above” when the phrase is the right answer & Percentage of ChatGPT's responses which are “(D) None/All of the Above” when the phrase is not the right answer \\ 
  \hline
  “None of the above” & 77 & 37\\ 
  \hline
  “All of the above” & 98 & 81\\ 
  \hline
\end{tabular}
\end{center}
\end{table}

In conclusion, creating questions with the answer choices “All of the above” or “None of the above”, but without the answer being those choices, is a very strong way to deter ChatGPT from correctly answering such questions. As shown before, ChatGPT’s accuracy in answering these types of questions is extremely low, because of its tendency to overpredict “(D)” in these cases. 

\section{Natural Language Processing Model}

We propose to build a tool that can answer whether a question will be answered accurately by ChatGPT, in order to help test makers to quickly identify and avoid test questions that are vulnerable to ChatGPT-based cheating. 
We made use of the Pytorch neural network framework and the  Google Colab GPU for training a model for finding the vulnerability of a question. 

\subsection{Preprocessing of Data}

Initially, we applied NLP preprocessing techniques to get each question into a more numerical format, which was easier to process for the neural network. This included tokenizing each question with the use of the BERT base model (uncased) \cite{tokenization}, which splits the text of the question up into words and word pieces (cases where one word can consist of multiple tokens). These tokens were then converted into token IDs, which are given by the BERT model. This process is demonstrated in Table 6. 

\begin{table}[htbp]
\caption{Word to Token IDs conversion}
\begin{center}
\begin{tabular}{ | p{1.5cm} | p{3cm} | p{2.5cm} |} 
  \hline
  \textbf{Word} & \textbf{Token(s)} & \textbf{Token ID(s)}\\ 
  \hline
  Reorganize & ["re", "\#\#org", "\#\#ani", "\#\#ze"] & [2128, 21759, 7088, 4371]\\ 
  \hline
  the & ["the"] & [1996]\\ 
  \hline
  hyper-parameters & ["hyper", "\#\#para", "\#\#meter", "\#\#s"] & [23760, 28689, 22828, 2015]\\ 
  \hline
  in & ["in"] & [1999]\\ 
  \hline
  the & ["the"] & [1996]\\ 
  \hline
  model. & ["model", "."] & [2944, 1012]\\ 
  \hline
\end{tabular}
\end{center}
\end {table}

From here, the data was eventually converted to PyTorch tensors and was ready for the neural network to train on.

We chose not to remove stop words, which are words that are so common that they are ignored by typical tokenizers. Some examples are "a", "the", "of", and "in". This technique was not used because it actually led to a decrease in accuracy for the model. 

\subsection{Dataset Splits}

Before training of the model started, we split the 11,657 questions and responses into a training dataset (75\%, used for the training model), a validation dataset (20\%, used for decreasing loss during training), and a testing dataset (5\%, used for determining accuracy of the final model). 

\subsection{Neural Network Architecture}

The model was instantiated with "bert-base-uncased", a pretrained BERT model. This model is pretrained on the English language and it uses a masked language modeling objective, meaning it tries to predict missing words in a sentence. 

The input into the model is first fed into twelve layers from the pretrained BERT model, which tokenizes and preprocesses the words from the problem. The hidden states (outputs) from these layers are fed into a BERT pooler layer, which uses a tanh() activation function. Finally, the hidden states from that layer are given to a dropout layer and a linear transformation is applied to that output before the final hidden states are received. This model is illustrated in Fig. 5.

\subsection{Training}

\begin{figure}
	   \includegraphics[width=0.5\textwidth]{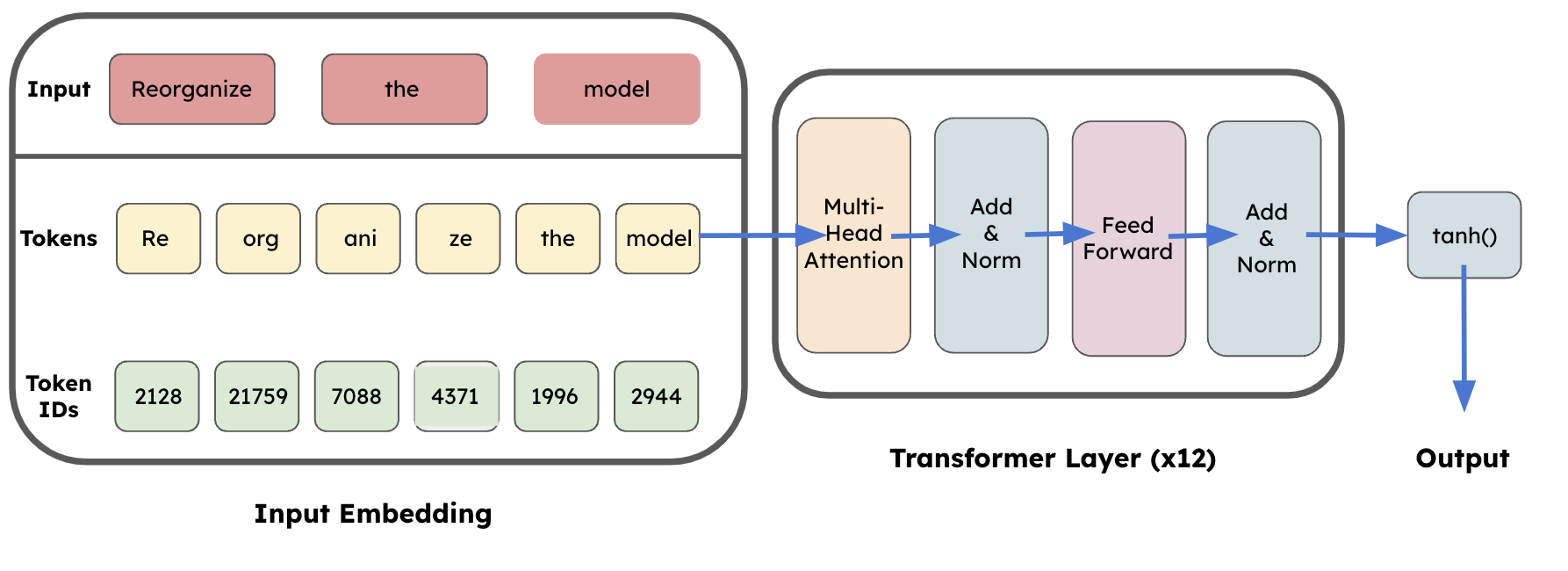}	
	\caption{Model Architecture} 
\end{figure}

We used the adam optimization algorithm \cite{adam} with a learning rate of $2e-5$ and an epsilon value of $1e-8$. The neural network was trained for 4 epochs. The model's highest validation accuracy was 0.6 and its lowest training loss was 0.52, as shown in Fig. 6. We ended up with a model that was correct in its answer of whether ChatGPT would correctly answer a question about 60\% of the time, which was satisfactory. 

\begin{figure}
	   \includegraphics[width=0.4\textwidth]{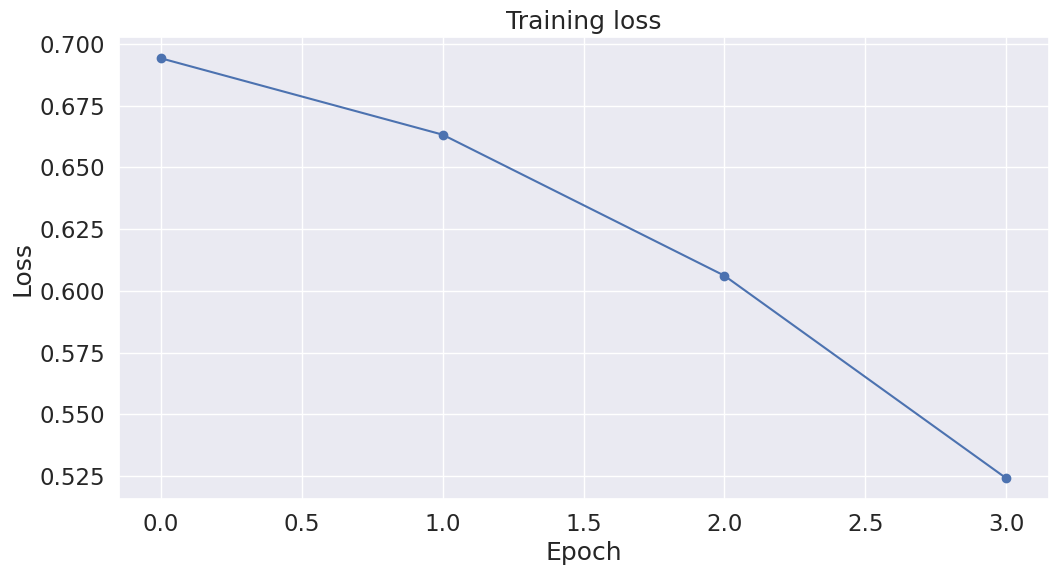}	
	\caption{Training loss of model} 
\end{figure}

\subsection{Other Models}

We have also tried other possible models, including Naïve Bayes Classifiers (NBC), decision trees \cite{tree}, random forests \cite{forest}, k-nearest neighbors (KNN), and logistic regression. For each of these models, we used the base model without adjusting any hyperparameters. Compared to the other models tested, neural networks provide the most proficient results in eliminating GPT-susceptible questions. The accuracies on the testing datasets for each of these models are shown in Table 7. 

\begin{table}[htbp]
\caption{Accuracy for different models}
\begin{center}
\begin{tabular}{ | p{3cm} | p{2cm} |} 
  \hline
  \textbf{Model} & \textbf{Accuracy} (\%) \\ 
  \hline
  Neural Network & 61\\ 
  \hline
  NBC & 47.4\\ 
  \hline
  Decision Tree & 57.3\\ 
  \hline
  Random Forest & 61.3\\ 
  \hline
  KNN & 55.7\\ 
  \hline
  Logistic Regression & 56.4\\ 
  \hline
\end{tabular}
\end{center}
\end{table}

\subsection{Testing of Model}

Although the model's accuracy was 60\%, when it was over 85\% certain in its answer, which happened about half of the time, its accuracy rose to over 70\%, which can filter out the majority of ChatGPT-vulnerable questions, given questions from the MedMCQA dataset.

We then proceeded to make a tool with the NLP model where an instructor can provide a certain set of questions and request to have a certain percentage of them removed, where the most susceptible questions to ChatGPT are taken out. Thus, we recorded the final accuracies based off of the testing dataset. The results are shown in Table 8.

\begin{table}[htbp]
\caption{Performance by final model}
\begin{center}
\begin{tabular}{ | p{4cm} | p{4cm} |} 
  \hline
  \textbf{Percentage of the questions to be removed} & \textbf{ChatGPT's accuracy on the remaining questions after removal} (\%) \\ 
  \hline
  0 & 61\\ 
  \hline
  5 & 60.2\\ 
  \hline
  10 & 58.9\\ 
  \hline
  15 & 57.9\\ 
  \hline
  20 & 56\\ 
  \hline
  30 & 54.7\\ 
  \hline
  40 & 53\\ 
  \hline
  50 & 53.6\\ 
  \hline
\end{tabular}
\end{center}
\end{table}

We find that ChatGPT's accuracy barely goes down when the percentage of questions removed goes up. However, we suspect when the model is trained more specifically on the exact type of questions being tested, this accuracy drop will increase.

\section{Alternate Dataset Testing}

In order to confirm the trends in ChatGPT’s answers discussed earlier, we gathered the same data on another smaller dataset. We compiled a dataset of physics problems since we wanted to test if our observations would hold true for a completely different type of problem. We decided on using a compilation of physics problems from a user on HuggingFace, and ended up with almost 500 physics problems to run through the Python API. On this dataset, ChatGPT’s accuracy was closer to 55\%, so the questions from the data challenged it more than the questions from the MedMCQA dataset. This can be attributed to the fact that physics problems typically require much more computation than the average medical school entrance exam question, and ChatGPT struggles more with computations. 

We started off by testing the observations we made with “None of the above” questions, where ChatGPT is much more likely to select “(D) None of the above” than any other options. In this case, when “D” was the right answer, the accuracy was 75\%, and when “D” was not the right answer, the accuracy was 50\%. This confirms our observations from before, proving ChatGPT’s tendency to overpredict “D”. The accuracies are shown in Table 9. 

\begin{table}[htbp]
\caption{Accuracy vs. presence of ”None of the above” option}
\begin{center}
\begin{tabular}{ | p{2.5cm} | p{2.5cm} | p{2.5cm} |} 
  \hline
  ChatGPT’s accuracy on questions where “None of the above” is the right answer & ChatGPT’s accuracy on questions where “None of the above” is not the right answer & ChatGPT’s accuracy on questions without “None of the above” as an answer choice \\ 
  \hline
  75\% & 50\% & 55\%\\ 
  \hline
\end{tabular}
\end{center}
\end{table}

Next, we tested the “All of the above” observations, where ChatGPT is much more likely to select “(D) All of the above” than any other option. In this case, when “D” was the right answer, the accuracy was 100\%, and when “D” was not the right answer, the accuracy was 27\%. Our observations were again confirmed, and like with the previous dataset, ChatGPT’s tendency to overpredict “D” is more exaggerated with “All of the above” questions. The accuracies are shown in Table 10. 

\begin{table}[htbp]
\caption{Accuracy vs. presence of ”All of the above” option}
\begin{center}
\begin{tabular}{ | p{2.5cm} | p{2.5cm} | p{2.5cm} |} 
  \hline
  ChatGPT’s accuracy on questions where “All of the above” is the right answer & ChatGPT’s accuracy on questions where “All of the above” is not the right answer & ChatGPT’s accuracy on questions without “All of the above” as an answer choice \\ 
  \hline
  100\% & 27\% & 55\%\\ 
  \hline
\end{tabular}
\end{center}
\end{table}

It must be noted that the data gathered from the second dataset comes from a very small sample. However, even though the results come from such a small sample, they can be trusted since they confirm observations coming from a much larger sample.

\section{Performance by Different Versions of GPT}

GPT-4 performs alarmingly better than any of the other models OpenAI has created. If someone were to use this model in an attempt to cheat, the likelihood of success would be substantially higher. However, most people use the gpt-3.5-turbo model, since that is the model being used on the main website for free. Additionally, GPT-4 is not free and consequently, a much lower percentage of people can afford to use it. The accuracies from different OpenAI models are shown in Table 11. 

\begin{table}[htbp]
\caption{Performance by different models of ChatGPT}
\begin{center}
\begin{tabular}{ | p{3cm} | p{3cm} |} 
  \hline
  Model & Accuracy (\%) \\ 
  \hline
  gpt-3.5-turbo & 60 \\ 
  \hline
  gpt-3.5-turbo-16k & 49 \\ 
  \hline
  text-davinci-003 & 47 \\ 
  \hline
  gpt-4 & 85 \\ 
  \hline
\end{tabular}
\end{center}
\end{table}

\section{Summary and conclusion}
Based on our results, we find that
test questions can be written in certain ways to prevent ChatGPT from answering them correctly. This includes the usage of certain phrases, such as "except", or the topic the questions test. There are also methods that can mistakenly be thought to fool ChatGPT, which do not, such as increasing question length, or using more complicated words. With the use of the techniques described in this paper, teachers can decrease the number of questions they put out that are susceptible to being answered correctly by ChatGPT. Additionally, they could attempt to filter out the GPT-susceptible questions in their tests by using an NLP model similar to the one we showed earlier.

In the future, we hope to test ChatGPT in other fields, such as math, computer science, and literature. We also hope to increase the accuracy of the current NLP model, such that it can filter out ChatGPT-susceptible questions with an even higher accuracy. Furthermore, we hope to validate or revise the trends noticed for ChatGPT for other chatbots such as Bard AI, Claude, or Bing AI. 

\section*{Acknowledgment}
C. Qian was partially supported by NSF Grants 1750704, 1932447, and 2114113.

\end{document}